\documentclass{svproc}
\usepackage{url}
\usepackage{graphicx,verbatim}
\usepackage{hyperref}
\usepackage{booktabs}
\usepackage{nicematrix}
\usepackage{multirow}
\usepackage{makecell}
\usepackage{subfig}
\usepackage{amssymb}
\usepackage{amsmath}
\usepackage[table]{xcolor}
\usepackage[separate-uncertainty = true,multi-part-units=single]{siunitx}

\begin{document}
\mainmatter              % start of a contribution
\title{Generalizing Soft Tissue Deformation and Force Prediction Across Material Stiffness and Geometry}
\titlerunning{Generalizing Soft Tissue Deformation}  % abbreviated title (for running head)
\author{Madina Kojanazarova \and Sidaty El Hadramy \and
Philippe C. Cattin}
\authorrunning{Madina Kojanazarova et al.} % abbreviated author list (for running head)
\tocauthor{Madina Kojanazarova, Sidaty El Hadramy, Philippe C. Cattin}
\institute{University of Basel, Department of Biomedical Engineering, Allschwil, Switzerland\\
\email{M.Kojanazarova@unibas.ch}}

\maketitle              % typeset the title of the contribution

\begin{abstract}
Accurate soft tissue simulation is essential for surgical training, pre-operative planning, and haptic feedback systems. 
While learning-based surrogate models trained on data using the finite element method (FEM) offer a promising path to real-time inference, their reliability depends on well-calibrated constitutive models.
Existing approaches neither provide systematic guidance on model selection across stiffness levels, nor generalize across different tissue stiffnesses or geometries.
We perform a comprehensive calibration of hyperelastic constitutive models in the SOFA Framework using gravity-loaded silicone beams with different stiffnesses. 
Using calibrated simulations as training data, we use a softness conditioned equivariant graph neural network, enabling deformation and force prediction across multiple tissue types and unseen geometries. 
Our model achieves sub-millimeter mean deformation accuracy at \qty{0.010}{\s} inference time, while showing that force prediction quality is directly tied to upstream calibration consistency.

\end{abstract}
\section{Introduction}
Minimally invasive surgery and robotic-assisted interventions increasingly rely on accurate models of soft tissue behavior to support surgical training, pre-operative planning, and intraoperative guidance~\cite{Cotin1999-sw}.
A core challenge in all of these applications is predicting how tissue deforms under tool contact in real time: deformation fields are needed to update augmented reality overlays~\cite{kenngott2014}, and contact forces must be rendered haptically to give trainees realistic tactile feedback~\cite{nguyen2023}.
Physics-based simulation using the finite element method (FEM) can reproduce soft tissue deformation with high fidelity by solving the equations of continuum mechanics on a volumetric mesh~\cite{Taylor2008-wk}, but the computational cost of solving large nonlinear systems at each time step typically precludes real-time use, with simulation times orders of magnitude above the millisecond budgets required for interactive applications~\cite{Nguyen2020}.

A widely adopted strategy to overcome this bottleneck is to use FEM offline to generate large bodies of high-fidelity training data, and then train a fast surrogate model that approximates the FEM response at inference time \cite{Mendizabal2020-gs,Deshpande2024-fy}.
Once trained, such surrogates can deliver predictions in milliseconds, enabling integration into real-time surgical simulators without sacrificing the physical realism of the underlying simulation. 
Among the available surrogate architectures, graph neural networks (GNNs) are particularly well-suited to this task.
They operate natively on unstructured meshes and point clouds, naturally handle irregular and patient-specific geometries, and can propagate local deformation cues globally through message passing \cite{ecgnn,Deshpande2024-fy}.

However, existing surrogates face key limitations. 
First, generalization across unseen geometries and varying material stiffness simultaneously remains an open challenge. 
Existing approaches typically fix one or both of these properties at training time~\cite{Deshpande2024-fy,El_Hadramy2025,kojanazarova2025,cGNN2025,Mendizabal2020-gs}. 
Second, any surrogate trained on FEM data is only as reliable as the underlying simulation. 
A critical and often overlooked prerequisite is the selection and calibration of the constitutive model~\cite{Hadramy2026latent,Mazier2022} and the mathematical description of how stress relates to strain in the tissue~\cite{Bertram2021-ig}. 
Simulation frameworks such as the SOFA Framework~\cite{sofa2012}, widely used in medical simulation research, provide several hyperelastic constitutive models, each encoding different assumptions about material microstructure. 
Yet there is no systematic guidance on which model best captures the behavior of silicone-like soft materials across different stiffness levels, and parameters overfit to a single configuration will directly corrupt the FEM training data and, by extension, the surrogate trained on it.

% Another limitation of existing GNN surrogates is the assumption of a fixed, known material stiffness. 
% Models are typically trained and evaluated on a single tissue formulation, limiting their applicability in settings where stiffness varies, soft robotics, and intraoperative tissue interaction, where different organs and pathological states exhibit substantially different mechanical properties \cite{Deshpande2024-fy,cGNN2025,kojanazarova2025,Mendizabal2020-gs}.

\begin{figure}%[b]
    \centering
    \includegraphics[width=\textwidth]{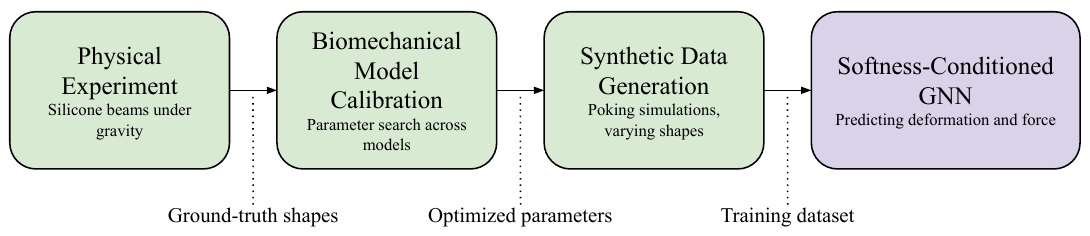}
    \caption{Overview of the proposed pipeline for experimental material calibration, synthetic data generation, and surrogate learning.}
    \label{methods_overview}
\end{figure}

% In this work, we address these limitations in a unified pipeline. 
% We fabricate silicone beams at three stiffness levels and use gravity-loading experiments across multiple lengths and boundary conditions as a controlled calibration benchmark. 
% We systematically optimize the parameters of all applicable hyperelastic models in SOFA \cite{sofa2012} and evaluate which models reproduce the physical deformation most accurately and which fail, particularly for very compliant materials. 
% Using the best-performing models and calibrated parameters, we generate a synthetic dataset of poking interactions on soft volumes with varied embedded geometries, spanning all three stiffness levels. 
% This dataset is used to train a softness-conditioned extension of the equivariant GNN of~\cite{kojanazarova2025}, enabling a single network to predict deformation and contact force across multiple material classes and generalize to unseen geometries, bringing us closer to a fast, generalizable surrogate for real-time soft tissue simulation.
In this work, we address these limitations in a unified pipeline, illustrated in Fig.~\ref{methods_overview}. 
First, we fabricate silicone beams at three stiffness levels and use gravity-loading experiments across multiple lengths and boundary conditions as a controlled calibration benchmark (Section~\ref{physical_setup}). 
We then systematically optimize the parameters of all applicable hyperelastic models in SOFA~\cite{sofa2012} and evaluate which models reproduce the physical deformation most accurately across stiffness levels (Section~\ref{param_opt}). 
Using the best-performing models and calibrated parameters, we generate a synthetic dataset of poking interactions on soft volumes with varied embedded geometries spanning all three stiffness levels (Section~\ref{data_gen}). 
Finally, this dataset is used to train a softness-conditioned extension of the equivariant GNN of~\cite{kojanazarova2025}, enabling a single network to predict deformation and contact force across multiple material classes and generalize to unseen geometries, bringing us closer to a fast, generalizable surrogate for real-time soft tissue simulation.

% Our pipeline consists of four stages, illustrated in Fig.~\ref{methods_overview}. 
% First, similar to \cite{Mazier2022}, we conduct physical gravity-loading experiments on silicone beams of three stiffness levels to obtain ground-truth deformation profiles (Section~\ref{physical_setup}). 
% Second, we replicate these experiments in the SOFA Framework, performing a systematic parameter search across all applicable hyperelastic constitutive models to identify which models best reproduce the observed physical behavior at each stiffness level (Section~\ref{param_opt}). 
% Third, using the best-performing models and their calibrated parameters, we generate a large synthetic dataset of poking interactions on soft volumes with embedded rigid inclusions of varying shape (Section~\ref{data_gen}). 
% Lastly, we use the generated dataset to train a softness-conditioned graph neural network model that predicts surface deformation and contact forces and generalizes across both unseen tissue stiffnesses and unseen shapes. 

\section{Background}
\subsection{Physics-Based Soft Tissue Simulations}
Soft tissues exhibit hyperelastic behavior where, under large deformations, the stress-strain relationship is nonlinear. 
This behavior is commonly modeled in continuum mechanics as a strain energy density function $\Psi$, defined as a function of the deformation gradient \textbf{F}, its determinant $J = \det \mathbf{F}$ (the local volume ratio), and the right Cauchy-Green tensor $\mathbf{C} = \mathbf{F}^T\mathbf{F}$, whose invariants are
\begin{equation}
I_1 = \text{tr}(\mathbf{C}), \quad I_2 = \tfrac{1}{2}\left[(\text{tr}\,\mathbf{C})^2 - \text{tr}(\mathbf{C}^2)\right].
\end{equation}
Hyperelastic models differ in how $\Psi$ is formulated, reflecting different assumptions about the underlying material microstructure and trade-offs between accuracy and numerical stability. 

\paragraph{Saint Venant-Kirchhoff} The simplest of the models considered, extends the classical linear (Hookean) stress-strain relationship to large displacements by replacing the linear strain tensor with the nonlinear Green-Lagrange strain tensor $\mathbf{E} = \tfrac{1}{2}(\mathbf{C} - \mathbf{I})$:
\begin{equation}
\Psi_{\text{StVK}} = \frac{\lambda}{2}\,\text{tr}(\mathbf{E})^2 + \mu\,\text{tr}(\mathbf{E}^2),
\end{equation}
with $\lambda$ and $\mu$ being the Lamé parameters. 
Because the stress remains a linear function of $\mathbf{E}$, the model is computationally efficient but physically unrealistic at large compressive strains, where it can produce non-physical, non-monotonic stress responses~\cite{Barbi2005}.

\paragraph{Neo-Hookean} Derived from the statistical mechanics of polymer chains under the assumption of Gaussian chain-length statistics, the compressible Neo-Hookean model is given by:
\begin{equation}
\Psi_{\text{NH}} = \frac{\mu}{2}(I_1 - 3) - \mu \ln J + \frac{\lambda}{2}(\ln J)^2.
\end{equation}
With only two material parameters, it is the simplest true hyperelastic model and captures the qualitative nonlinearity of rubber-like materials, but it does not reproduce the pronounced strain-stiffening observed in real elastomers at large stretches~\cite{Mihai2015}.

\paragraph{Stable Neo-Hookean} A more recent reformulation of the Neo-Hookean model designed for robustness in simulation rather than fidelity to a specific microstructural derivation:
\begin{equation}
\Psi_{\text{SNH}} = \frac{\mu}{2}(I_1 - 3) - \mu(J - 1) + \frac{\lambda+\mu}{2}(J-1)^2.
\end{equation}
By modifying the volumetric term relative to the standard compressible Neo-Hookean model, this formulation guarantees a positive semi-definite Hessian even under element inversion or extreme compression, eliminating the numerical instabilities that can otherwise arise during large or fast deformations~\cite{Smith2018}.

\paragraph{Mooney-Rivlin} Proposed as a generalization of the Neo-Hookean model. The Mooney-Rivlin model introduces a second invariant term to better fit experimental stress-strain data over a wider strain range:
\begin{equation}
\Psi_{\text{MR}} = C_{10}(\bar{I}_1 - 3) + C_{01}\bar{I}_2 + \frac{K}{2}(\ln J)^2,
\end{equation}
where $\bar{I}_1$, $\bar{I}_2$ are the volume-normalized invariants and $K$ is the bulk modulus. 
The Neo-Hookean model is recovered as the special case when $C_{01}=0$~\cite{Mihai2015}.

\paragraph{Ogden} Rather than expressing $\Psi$ in terms of invariants, the Ogden model is formulated directly in terms of the principal stretches $\lambda_1$, $\lambda_2$, $\lambda_3$:
\begin{equation}
\Psi_{\text{Ogden}} = \sum_{p=1}^{N} \frac{\mu_p}{\alpha_p}\left(\lambda_1^{\alpha_p} + \lambda_2^{\alpha_p} + \lambda_3^{\alpha_p} - 3\right) + \frac{K}{2}(\ln J)^2.
\end{equation}
This formulation offers greater flexibility in fitting complex, large-strain stress-stretch behavior, but with the need for an eigen-decomposition of the deformation tensor at each step, increasing computational cost~\cite{Kim2012}.

\paragraph{Arruda-Boyce} Motivated by a non-Gaussian statistical model of polymer chains arranged on an eight-chain unit cell, the Arruda-Boyce model captures the limiting chain extensibility responsible for strain-stiffening at large stretches:
\begin{equation}
\Psi_{\text{AB}} = \mu \sum_{i=1}^{5} \frac{C_i}{\lambda_m^{2i-2}} \left(I_1^i - 3^i\right) + \frac{K}{2}(\ln J)^2,
\end{equation}
where $C_1, \dots, C_5$ are fixed coefficients and $\lambda_m$ is the locking stretch, the stretch at which the polymer network becomes fully extended, and it captures strain-stiffening behavior that the Neo-Hookean and Mooney-Rivlin models cannot~\cite{ARRUDA1993}.

\subsection{Graph Neural Network for Soft Tissue Simulation}
Graph neural networks propagate information through message-passing layers that update node coordinates and features based on local neighborhoods, making them well-suited for unstructured mesh and point cloud data~\cite{Hamilton2020-sf}. 
Equivariance to rotations and translations allows the network to generalize across orientations without explicit data augmentation~\cite{ecgnn}. 
The conditional equivariant GNN of~\cite{kojanazarova2025} builds on this principle, taking as input a surface point cloud \( \mathbf{x}^0 \in \mathbb{R}^{N \times d} \) ($N$ surface points in 3D), per-point tissue features \( \mathbf{h} \in \mathbb{R}^{N \times d} \) encoding subsurface material composition sampled at 128 depth intervals beneath each point, and condition vectors $\mathbf{C}_s, \mathbf{C}_e \in \mathbb{R}^3$ representing the start and end coordinates of the applied force.
% condition feature vector \( \mathbf{C}_h \in \mathbb{R}^{4 \times 128} \), which encodes the binary tissue profile and corresponding 3D coordinates sampled along the straight line from \( \mathbf{C}_s \) to \( \mathbf{C}_e \), then flattened into a single vector
Information is propagated through $L$ equivariant message-passing layers, and predictions are produced by two dedicated branches: one regressing per-point displacements \( \mathbf{u} \in \mathbb{R}^{N \times 3} \) and one estimating scalar contact force $F$, with the deformed point cloud recovered as \( \hat{\mathbf{y}} = \mathbf{u} + \mathbf{x^0} \).
We adopt this architecture without modification; full details are given in~\cite{kojanazarova2025}. The sole adaptation concerns the tissue feature encoding $\mathbf{h}$, described in Section~\ref{training}.

% The input to the model is a point cloud \( \mathbf{x}^0 \in \mathbb{R}^{N \times d} \), where \( N \) is the number of surface points and \( d = 3 \) corresponds to their 3D coordinates. 
% Each point has tissue features \( \mathbf{h} \in \mathbb{R}^{N \times d_{h}} \), where \( d_{h} = 256 \) encodes a flattened vector representing the binary tissue values (soft or hard) and their corresponding z-axis positions sampled beneath each point. 

% The input also includes a conditional vector $\mathbf{C}_s, \mathbf{C}_e \in \mathbb{R}^3$ representing the 3D start and end coordinates of the applied force, and a condition feature vector \( \mathbf{C}_h \in \mathbb{R}^{4 \times 128} \), which encodes the binary tissue profile and corresponding 3D coordinates sampled along the straight line from \( \mathbf{C}_s \) to \( \mathbf{C}_e \), then flattened into a single vector.

% The output of the displacement prediction branch of the model is the predicted displacement \( \mathbf{u} \in \mathbb{R}^{N \times 3} \), which we apply to the input points to obtain the deformed point cloud \( \hat{\mathbf{y}} = \mathbf{u} + \mathbf{x^0} \), and the output of the force branch is the force $F$.

% We adopt this architecture as our backbone without modification; full details are given in~\cite{kojanazarova2025}. 
% The sole adaptation introduced in this work concerns the tissue feature encoding, described in Section~\ref{training}.

\section{Methods}
\subsection{Physical Setup}\label{physical_setup}
Silicone beams of three different stiffnesses were fabricated by varying the softener concentration in the mixture. 
Each beam was cast using a two-part silicone rubber (Eurosil 4 $A+B$, Schouten SynTec, the Netherlands) combined with an oil-based softener (Eurosil Softener, Schouten SynTec, the Netherlands), mixed in an $A$:$B$:$S$ ratio by weight,
with $A$:$B$ fixed at 1:1 and $S$ varied at 0, 0.5, and 1. 
Throughout this paper, each mixture is referred to by its softener ratio, e.g. $S0$, $S0.5$, and $S1$ (corresponding to A:B:S = 1:1:0, 1:1:0.5, and 1:1:1).

\begin{figure}[!b]
    \centering
    \includegraphics[width=\textwidth]{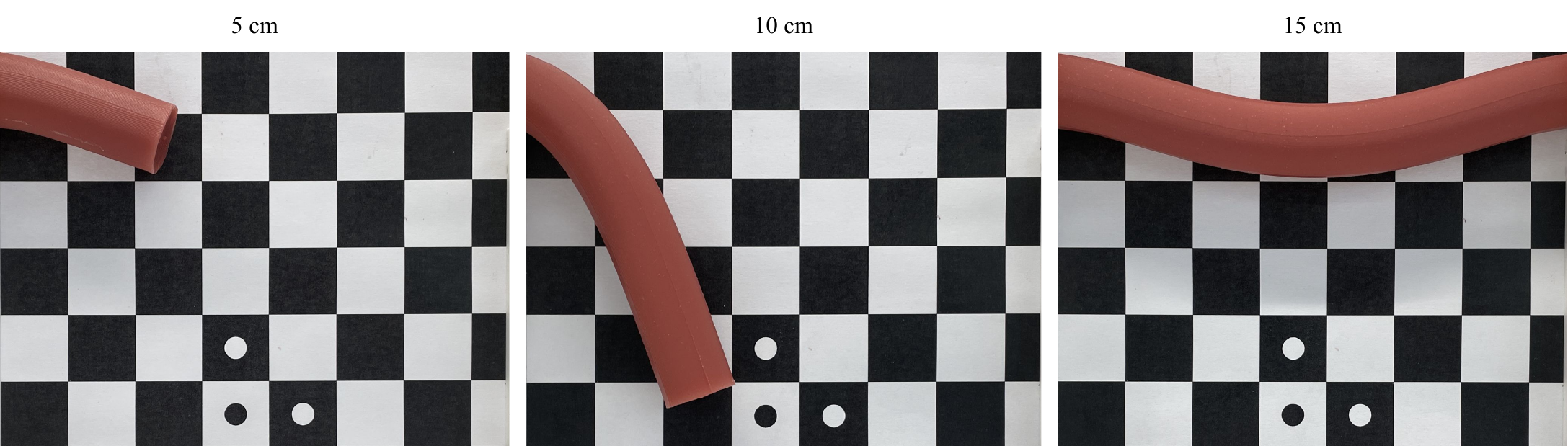}
    \caption{Silicone beams at different lengths for the softener ratio $S0.5$.}
    \label{F_beams}
\end{figure}

The cylindrical molds were 3D-printed with a length of \qty{200}{\mm} and a radius of \qty{10}{\mm}. 
Due to minor silicone leakage during casting, the beams varied slightly in final length. 
To obtain consistent and comparable deflection measurements, each beam was tested at standardized lengths: \qty{50}{\mm} and \qty{100}{\mm} in a cantilevered configuration (clamped at one end, free at the other), and \qty{150}{\mm} in a clamped-clamped (fixed-fixed) configuration. 
In each case, deformation was recorded after the beam had fully stabilized under gravity. 
% The deformed profiles were captured via imaging (Fig.~\ref{F_beams}). The resulting images were corrected for lens distortion, segmented and binarized to extract the beam centerline. 
% The ground-truth deformed geometry was then reconstructed in 3D using CAD, serving as the reference for parameter calibration.
The deformed profiles were captured using a smartphone camera with a checkerboard reference pattern included in each frame (Fig.~\ref{F_beams}). 
We used OpenCV to compute a homography correcting perspective distortion and establishing an image resolution of \qty{0.086 \pm 0.006}{mm/pixel}. 
The corrected images were imported into CAD software and scaled using both the reference grid and the known diameter of an undeformed beam, minimizing scaling inaccuracies and compensating for perspective error from camera-to-specimen distance variation.
The outer contour of the deformed specimen was manually traced from the calibrated images, and a centerline corresponding to the neutral axis was generated from this contour using the CAD software's centerline function.
The known cross-sectional diameter was then swept along this centerline, constrained by the traced outer contour, to reconstruct the 3D deformed geometry, serving as the reference for parameter calibration.

\begin{figure}%[!t]
    \centering
    \includegraphics[width=\textwidth]{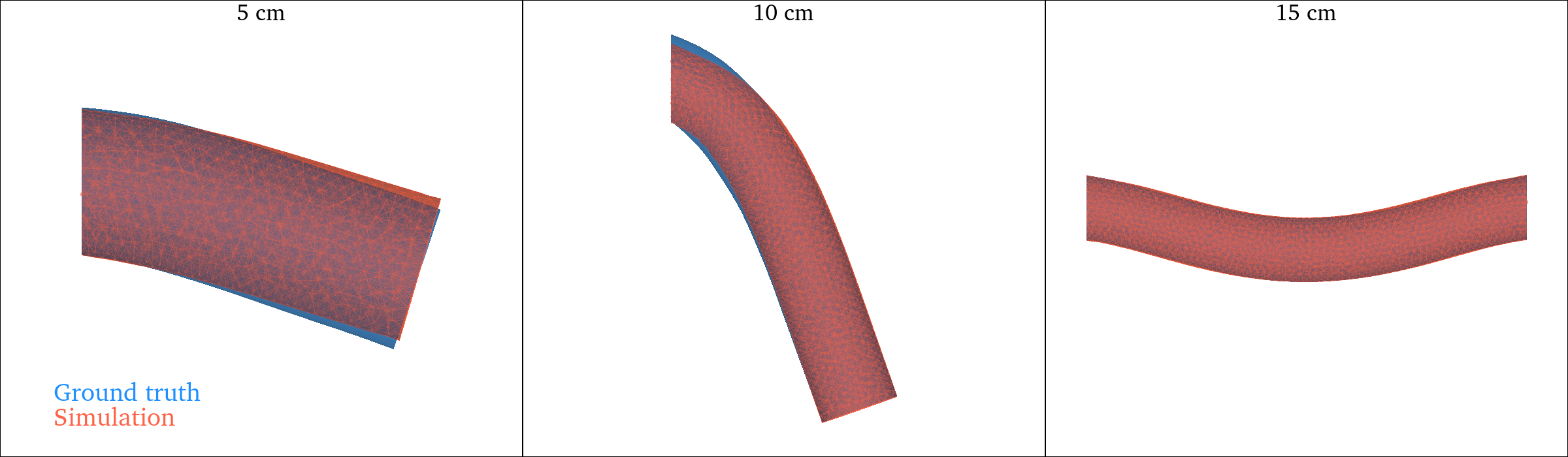}
    \caption{Optimization results for the Ogden model for softness $S0.5$.}
    \label{opt_og}
\end{figure}

\subsection{Biomechanical Model \& Parameter Optimization}\label{param_opt}
For each silicone mixture, we simulated the gravity-induced deformation of a cylindrical mesh with $7257$ nodes.
% defining the boundary condition to match the dimensions of the physical mold. 
The beam was fixed in SOFA, setting the boundary condition at the ends of the beam to match the physical experiments (Section~\ref{physical_setup}). 
We rigidly constrained all translational degrees of freedom of every mesh node within a finite-length region at each clamped end (rather than a single end cross-section), matching the finite grip length of a physical clamp; no rotation or sliding was permitted within the clamped regions. The \qty{50}{} and \qty{100}{\mm} beams were clamped at one end only (cantilevered, free at the other), while the \qty{150}{\mm} beam was clamped identically at both ends.
The mass density was set to \qty{1000}{kg/m^3} for all beams as the density variation due to softener content was considered negligible.

We performed a parameter search over all hyperelastic constitutive models available in SOFA that are applicable to incompressible or near-incompressible soft materials. Each SOFA simulation ran for up to \num{10000} steps at a \qty{0.001}{s} time step. 
The simulation was run until equilibrium, defined as the root-mean-square (RMS) nodal displacement, measured as the change in node position over consecutive 50-step windows, remaining below \qty{1}{\mm}, averaged across three consecutive windows (\num{150} steps total).
For each model and each softener ratio $S$, parameters were optimized by maximizing the average volumetric overlap error across all three beam lengths, measured using the Dice score between the simulated and ground-truth 3D surface meshes. 
Optimizing across lengths simultaneously ensures the recovered parameters are consistent across boundary conditions rather than tuned to a single configuration.
The optimization results and metrics for each model and softness level are reported in Table~\ref{beam_opt}.
We observe that most models could replicate the beam behavior with Dice scores $>0.90$ for the $S0$ and $S0.5$ softnesses. 
Although the Ogden model was the slowest, it outperforms the other models in the $S1$ case and manages to reflect the softer tissue best. 
Based on these results, Ogden and Mooney-Rivlin were selected for data generation, as they achieved the most reliable fits across all three softness levels.

\begin{table}[!t]
\centering
\setlength{\tabcolsep}{0.3em}
\renewcommand{\arraystretch}{0.85}
\caption{Hyperelastic constitutive model calibration results across three softness levels ($S0$, $S0.5$, $S1$), optimized by maximizing volumetric overlap (Dice) between simulated and ground-truth beam deformations. For each softness level, metrics are averaged over the three tested beam lengths (\qty{50}{\mm}, \qty{100}{\mm}, and \qty{150}{\mm}). Surface distance, 95th percentile Hausdorff distance (HD95), Dice score, volume difference, and simulation time are reported per softness level and as an average.}\label{beam_opt}
\begin{tabular}{ll|ccccc}
    \makecell[tl]{Model}  
    & \makecell[tl]{Softness} 
    & \makecell{Surface\\dist$\downarrow$ [mm]} 
    & \makecell{HD95$\downarrow$ [mm]} 
    & \makecell{Dice$\uparrow$}
    & \makecell{Vol.\\diff.$\downarrow$ [\%]} 
    & \makecell{Time$\downarrow$ [s]} \\
    \midrule
    \multirow{4}{*}{\shortstack[l]{Arruda-\\Boyce}} & $S0$ & $1.697$ & $4.333$ & $0.955$ & $0.732$ & $125.70$\\ 
    & $S0.5$ & $1.765$ & $4.521$ & $0.936$ & $2.606$ & $83.04$\\ 
    & $S1$ & $10.281$ & $22.661$ & $0.571$ & $2.802$ & $92.30$ \\
    & Average & $4.581$ & $10.505$ & $0.821$ & $2.047$ & $100.35$\\
    % & :$1.5S$ & $1.740$ & $4.440$ & $0.941$ & $2.510$ \\ 
    \midrule
    \multirow{4}{*}{\shortstack[l]{Mooney-\\Rivlin}} & $S0$ & $1.706$ & $4.352$ & $0.953$ & $0.731$ & $66.09$\\ 
    & $S0.5$ & $1.753$ & $4.502$ & $0.943$ & $2.605$ & $42.18$\\ 
    & $S1$ & $2.008$ & $5.074$ & $0.889$ & $2.274$ & $41.62$\\ 
    & Average & $1.822$ & $4.643$ & $0.923$ & $1.870$ & $49.96$ \\ 
    % & :$1.5S$ & $1.849$ & $4.850$ & $0.914$ & $2.528$ \\ 
    \midrule
    \multirow{4}{*}{Neo-Hookean} & $S0$ & $1.691$ & $4.330$ & $0.957$ & $0.735$ & $47.29$\\ 
    & $S0.5$ & $1.838$ & $4.792$ & $0.916$ & $2.617$ & $ 34.09$\\ 
    & $S1$ & $1.993$ & $5.032$ & $0.891$ & $2.302$ & $26.89$\\ 
    & Average & $1.841$ & $4.718$ & $0.921$ & $1.885$ & $36.09$\\ 
    % & :$1.5S$ & $1.824$ & $4.556$ & $0.915$ & $2.544$ \\ 
    \midrule
    \multirow{4}{*}{\shortstack[l]{Stable\\Neo-Hookean}} & $S0$ & $1.718$ & $4.349$ & $0.951$ & $0.738$ & $45.06$\\ 
    & $S0.5$ & $1.842$ & $4.727$ & $0.914$ & $2.625$ & $28.50$\\ 
    & $S1$ & $3.474$ & $9.625$ & $0.728$ & $2.232$ & $30.40$\\ 
    & Average & $2.345$ & $6.234$ & $0.868$ & \textbf{1.865} & $34.65$\\ 
    % & :$1.5S$ & $1.868$ & $4.758$ & $0.901$ & $2.598$ \\ 
    \midrule
    \multirow{4}{*}{Ogden} & $S0$ & $1.773$ & $4.420$ & $0.934$ & $0.710$ & $245.48$\\ 
    & $S0.5$ & $1.677$ & $4.380$ & $0.967$ & $2.583$ & $220.20$\\ 
    & $S1$ & $1.877$ & $4.822$ & $0.926$ & $4.068$ & $ 153.31$\\
    & Average & \textbf{1.776} & \textbf{4.541} & \textbf{0.942} & $2.454$ & $203.33$\\
    % & :$1.5S$ & $1.768$ & $4.607$ & $0.933$ & $2.563$ \\
    \midrule
    \multirow{4}{*}{\shortstack[l]{St Venant-\\Kirchhoff}} & $S0$ & $1.700$ & $4.319$ & $0.952$ & $0.948$ & $20.12$\\ 
    & $S0.5$ & $1.994$ & $5.054$ & $0.872$ & $3.507$ & $35.89$\\ 
    & $S1$ & $2.631$ & $6.696$ & $0.739$ & $3.628$ & $18.82$\\ 
    & Average & $2.108$ & $5.356$ & $0.854$ & $2.694$ & \textbf{24.95}\\ 
    % & :$1.5S$ & $2.383$ & $6.073$ & $0.778$ & $5.323$ \\ 
    % \botrule
\end{tabular}
\end{table}

\subsection{Data Generation}\label{data_gen}
\begin{figure}[!t]
    \centering
    \includegraphics[width=\textwidth]{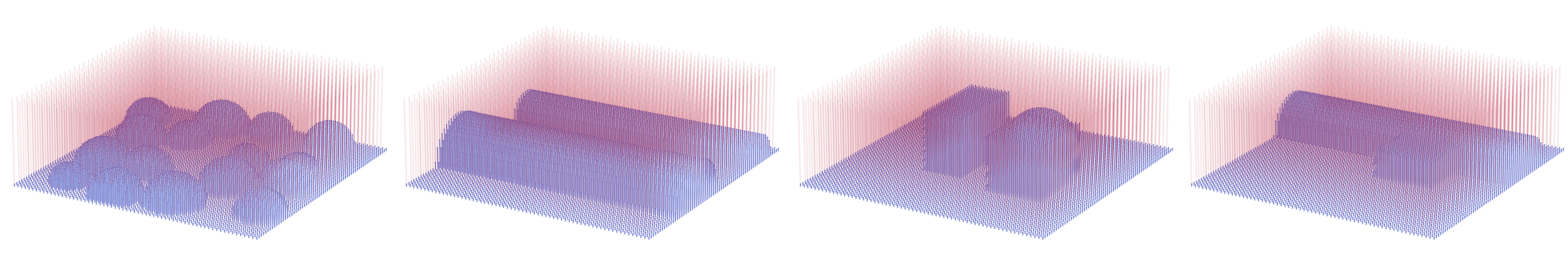}
    \caption{Examples of generated binary shapes of soft tissues with rigid inclusions.}
    \label{shapes}
\end{figure}

Training data was generated using the two best-performing models from the parameter optimization study, Ogden and Mooney-Rivlin, applied across all three softness levels. 
We constructed a rectangular soft tissue volume of \qtyproduct[product-units=power]{100 x 100 x 32}{\mm} with an embedded rigid inclusion.
Eleven distinct inclusion shapes were designed, comprising cubes, spheres and cylinders of varying sizes, randomly positioned at the base of the volume (Fig.~\ref{shapes}). 
The inclusion surfaces and the bottom face together defined the rigid boundary, while the remaining volume was discretized into tetrahedral elements representing the soft tissue.

For each shape, we sampled two sets of poking locations on the top surface: $20$ fixed locations shared across all three softness levels, enabling direct comparison of the tissue response to softness alone, and $20$ randomly sampled locations drawn independently per softness level to increase contact configuration diversity 
In both cases, the poking angle was drawn up to 45° from the surface normal, with a maximum penetration depth of \qty{30}{\mm}.

The mesh was locally refined around each poking site and poking was performed using a cylindrical rod with a radius of \qty{2.5}{\mm} and length of \qty{40}{\mm}.
Mesh resolution was determined via a convergence study on a homogeneous volume without rigid inclusion, comparing output forces at a penetration depth of \qty{10}{\mm}, chosen because not all models are guaranteed to reach the full \qty{30}{\mm} across all softness levels. The resulting dataset meshes ranged from \num{13297} to \num{17732} nodes (mean \num{15954} $\pm$ \num{1254}) and \num{58330} to \num{86130} elements (mean \num{75792} $\pm$ \num{8116}) across all shapes and softness levels.

Each poking simulation was prescribed over \qty{2}{\s} with a time step of \qty{0.002}{\s}. 
The recorded outputs at each time step were the positions of all surface nodes and the total reaction force at the boundary nodes. 
To remove near-static timesteps, frames were retained only when the surface displacement at the contact tip exceeded \qty{1}{\mm} relative to the previous retained frame, yielding a dataset of $11\times3\times40=1320$ pokings, with $14305$ snapshots in the Ogden dataset and $13672$ snapshots in the Mooney-Rivlin dataset.
The complete features of the generated dataset can be seen in Table~\ref{data_features}.

\begin{table}[!b]
\centering
\setlength{\tabcolsep}{0.5em}
\caption{Features of the simulated dataset. $Sim_{max}$ and $F_{max}$ indicate the average maximum depth and force reached per poke. $Tip_{mean}$ and $F_{mean}$ indicates the average depth and force in each snapshot.}\label{data_features}
\begin{tabular}{ll|cccc}
    \makecell[tl]{Model}  
    & \makecell[tl]{Softness} 
    & \makecell{$Sim_{max}$ [mm]} 
    & \makecell{$F_{max}$ [N]}
    & \makecell{$Tip_{mean}$ [mm]} 
    & \makecell{$F_{mean}$ [N]} \\
    \midrule
    \multirow{3}{*}{\shortstack[l]{Mooney-\\Rivlin}} & $S0$ & $13.92\pm3.88$ & $12.92\pm5.32$ & $7.91\pm4.82$ & $4.80\pm4.59$ \\ 
    & $S0.5$ & $13.78\pm4.14$ & $6.19\pm2.69$  & $7.94\pm4.89$& $2.33\pm2.30$ \\ 
    & $S1$ & $13.46\pm4.24$ & $3.87\pm2.43$ & $7.82\pm4.87$ & $1.42\pm1.54$ \\ 
    \midrule
    \multirow{3}{*}{Ogden} & $S0$ & $16.32\pm6.53$ & $29.71\pm25.31$ & $9.88\pm6.54$ & $6.94\pm11.44$ \\ 
    & $S0.5$ & $16.40\pm6.41$ & $18.35\pm18.67$ & $9.89\pm6.54$ & $3.94\pm7.47$ \\ 
    & $S1$ & $11.20\pm4.82$ & $2.34\pm1.98$ & $7.04\pm4.87$ & $1.01\pm1.19$ \\ 
    % \botrule
\end{tabular}
\end{table}

\subsection{Training Details}\label{training}
We sampled $32 \times 32$ surface points on a regular grid from the FEM simulations to use as the input point cloud. 
In the original work \cite{kojanazarova2025}, the depth profiles \(\mathbf{h}\) were assigned a binary label: 0 for soft tissue and 1 for bone, reflecting the binary nature of the tissue. 
Since our dataset spans three silicone mixtures with distinct stiffnesses, we assigned each material a continuous stiffness value $s\in[0,1]$, where higher values indicate stiffer material. 
The rigid inclusions retain a value of $1.0$, while each silicone formulation is assigned a value derived from its softener ratio $S$, normalized such that the stiffest silicone $S0$ maps to $0.8$, $S0.5$ maps to $0.4$ and the softest $S1$ maps to $0.0$.
We split the data into training:validation:test in the order of shapes in an 7:2:2 ratio with 5-fold cross-validation, and trained the two datasets (Ogden and Mooney-Rivlin) separately for comparison, both including all softnesses.

The models were trained for $200$ epochs using the Adam optimizer with a learning rate initialized to $10^{-4}$ and a batch size of $64$.
The loss function was defined the same as in \cite{kojanazarova2025}, minimizing the weighted Euclidean distance error $\mathcal{L}_d$ and MSE for the force predictions $\mathcal{L}_f$. 
We report all deformation errors computed using the standard (unweighted) Euclidean distances.

The FEM simulations were performed on an AMD Epyc 7742 64-core CPU, and the GNN models were trained in PyTorch using the PyTorch Geometric (v2.8.0) package and run on a single NVIDIA RTX 2080 GPU.

\begin{table}[!b]
\centering
\setlength{\tabcolsep}{0.5em}
\caption{GNN prediction results for the different datasets and softness levels.}\label{results}
\begin{tabular}{ll|cccc}
    \makecell[tl]{Dataset}
    & \makecell[tl]{Softness} 
    & \makecell{Force\\Absolute\\Error [N]} 
    & \makecell{Mean\\Euclidean\\Distance [mm]} 
    & \makecell{Max\\Euclidean\\Distance [mm]}
    & \makecell{Mean\\ Relative Tip\\Error [\%]} \\
    \midrule
    \multirow{4}{*}{\shortstack[l]{Mooney-\\Rivlin}} & $S0$ & $0.537\pm1.001$ & $0.097\pm0.152$ & $1.205\pm0.854$ & $2.3\pm2.5$ \\ 
    & $S0.5$ & $0.310\pm0.540$ & $0.097\pm0.155$ & $1.228\pm0.924$ & $2.1\pm2.3$ \\ 
    & $S1$ & $0.258\pm0.650$ & $0.092\pm0.157$ & $1.268\pm0.944$ & $2.2\pm2.6$ \\ 
    & Average & $0.337\pm0.708$ & $0.095\pm0.155$ & $1.239\pm0.918$ & $2.2\pm2.4$ \\ 
    \midrule
    \multirow{3}{*}{Ogden} & $S0$ & $2.269\pm6.358$ & $0.127\pm0.209$ & $1.496\pm1.084$ & $2.5\pm2.6$ \\ 
    & $S0.5$ & $1.407\pm5.155$ & $0.132\pm0.219$ & $1.517\pm1.104$ & $2.4\pm2.5$ \\ 
    & $S1$ & $0.222\pm0.776$ & $0.079\pm0.145$ & $1.237\pm1.013$ & $2.1\pm2.7$ \\ 
    & Average & $1.253\pm4.751$ & $0.114\pm0.198$ & $1.427\pm1.080$ & $2.3\pm2.6$ \\ 
    % \botrule
\end{tabular}
\end{table}

\section{Results \& Discussion}
Table~\ref{results} reports deformation and force prediction errors per softness level and per constitutive model used to generate the training data (Mooney-Rivlin and Ogden), alongside the total across all conditions.

\paragraph{Deformation accuracy} 
Mean Euclidean distances are low and consistent across all conditions, averaging \qty{0.095\pm0.155}{\mm} for Mooney-Rivlin and \qty{0.114\pm0.198}{\mm} for Ogden, both below the \qty{0.156\pm0.298}{\mm} reported in~\cite{kojanazarova2025} for a single-material and shape model.
This suggests that extending training to multiple softness levels, combined with the continuous stiffness encoding, does not degrade deformation accuracy relative to the single-material baseline. The lower errors are likely due to the increased diversity of training configurations and shapes. 
% This comparison should be interpreted with care as the lower errors may partly reflect dataset differences notably that~\cite{kojanazarova2025} trained on a single phantom geometry, and had deeper deformations compared to our Mooney-Rivlin dataset
The mean relative tip error remains consistently around $2-2.5$\% across all conditions for both models, further confirming reliable deformation prediction across the stiffness range.

\paragraph{Force prediction} 
Force errors are more variable and show a clear dependence on both the constitutive model and softness level. 
For Mooney-Rivlin, force errors are low and consistent across softness levels (\qty{0.258}{\N} to \qty{0.537}{\N}), compared to those reported in~\cite{kojanazarova2025} (\qty{0.701\pm1.111}{\N}). 
For Ogden, force errors are substantially higher and more variable for the stiffer mixture ($S0$: \qty{2.269\pm6.358}{\N}, before dropping sharply for the softest level ($S1$: \qty{0.222\pm0.776}{\N}). 
This pattern directly reflects the characteristics of the training data: as seen in the data features in Table~\ref{data_features}, the Ogden model produces large and highly variable forces for stiffer materials, particularly $S0$ and $S0.5$, where the simulation reaches greater penetration depths before stabilizing. 
The resulting high variance in training targets makes force regression harder, and more training examples or a stronger tissue stiffness encoding in the model would likely reduce these errors.
This high variance also partly reflects the imperfect calibration fit for stiffer Ogden conditions: parameters that do not perfectly reproduce physical behavior produce inconsistent force outputs across similar configurations, which directly increases the difficulty of the regression task.

\paragraph{Effect of constitutive model} 
The Mooney-Rivlin dataset produces more consistent GNN predictions overall, which we attribute to its more uniform force distributions across softness levels rather than superior physical accuracy. 
The Mooney-Rivlin simulations reach shallower maximum depths on average (Table~\ref{data_features}), which limits force variability and results in a more learnable dataset. 
Ogden, by contrast, sustains larger deformations and generates physically richer force profiles, particularly for stiffer tissues, but at the cost of higher variance that the current model struggles to fit. 
This distinction is practically important, as Mooney-Rivlin may be preferable when training data consistency is the priority, while Ogden may be more appropriate when the goal is to cover the full range of tissue responses, provided sufficient training data and a more expressive tissue encoding are available. 
In both cases, deformation accuracy remains comparable, suggesting that the GNN's displacement prediction is robust to the choice of constitutive model, while force prediction is more sensitive to training data characteristics.

Across both datasets, a single inference took \qty{0.010\pm0.004}{\s} (\qty{100}{\Hz}), providing close to real-time predictions required for interactive surgical simulations.
% , and slightly faster than the \qty{0.015\pm0.001}{\s} reported in~\cite{kojanazarova2025}, likely attributable to updates in the PyTorch Geometric library rather than architectural differences.
Additionally, as demonstrated in~\cite{kojanazarova2025}, the underlying architecture generalizes across point cloud densities and structures, a property inherited by our extension without additional training.

\section{Conclusions}
We presented a pipeline for generalizable soft tissue simulation combining systematic calibration of SOFA hyperelastic constitutive models with a softness-conditioned graph neural network surrogate. 
Our calibration study identified Ogden and Mooney-Rivlin as the most reliable models across tested stiffness levels, while revealing the limitations of current constitutive models for very compliant materials. 
Conditioning the GNN on material stiffness enabled accurate deformation and force across three trained material softness levels and unseen inclusion geometries, maintaining sub-millimeter mean deformation accuracy at close to real-time inference speeds.
These findings support the use of carefully calibrated FEM simulations as a basis for fast, generalizable data-driven surrogates in surgical simulation. 
The close to real-time inference speed of \qty{0.010}{\s} demonstrates the practical viability of such surrogates for integration into interactive surgical training systems with haptic feedback.
% Future work should investigate more physically grounded stiffness encodings, for instance derived from calibrated material parameters rather than softener ratios — and explore improved constitutive models for very compliant materials to further close the gap between simulation and physical tissue behavior.
\bibliographystyle{spmpsci}
\bibliography{mybibliography}
% \begin{thebibliography}{6}
% %

% \bibitem {smit:wat}
% Smith, T.F., Waterman, M.S.: Identification of common molecular subsequences.
% J. Mol. Biol. 147, 195?197 (1981). \url{doi:10.1016/0022-2836(81)90087-5}

% \bibitem {may:ehr:stein}
% May, P., Ehrlich, H.-C., Steinke, T.: ZIB structure prediction pipeline:
% composing a complex biological workflow through web services.
% In: Nagel, W.E., Walter, W.V., Lehner, W. (eds.) Euro-Par 2006.
% LNCS, vol. 4128, pp. 1148?1158. Springer, Heidelberg (2006).
% \url{doi:10.1007/11823285_121}

% \bibitem {fost:kes}
% Foster, I., Kesselman, C.: The Grid: Blueprint for a New Computing Infrastructure.
% Morgan Kaufmann, San Francisco (1999)

% \bibitem {czaj:fitz}
% Czajkowski, K., Fitzgerald, S., Foster, I., Kesselman, C.: Grid information services
% for distributed resource sharing. In: 10th IEEE International Symposium
% on High Performance Distributed Computing, pp. 181?184. IEEE Press, New York (2001).
% \url{doi: 10.1109/HPDC.2001.945188}

% \bibitem {fo:kes:nic:tue}
% Foster, I., Kesselman, C., Nick, J., Tuecke, S.: The physiology of the grid: an open grid services architecture for distributed systems integration. Technical report, Global Grid
% Forum (2002)

% \bibitem {onlyurl}
% National Center for Biotechnology Information. \url{http://www.ncbi.nlm.nih.gov}

% \end{thebibliography}
\end{document}